\documentclass[10pt,twocolumn,letterpaper]{article}

\usepackage{wacv}
\usepackage{times}
\usepackage{epsfig}
\usepackage{graphicx}
\usepackage{amsmath}
\usepackage{amssymb}

\usepackage{color}
\definecolor{note}{rgb}{0.1,0.1,1}

\usepackage{enumitem}
\usepackage{pgfplots}
\usepackage{tikz,pgfplots}

\usepackage[pagebackref=true,breaklinks=true,letterpaper=true,colorlinks,bookmarks=false]{hyperref}

\wacvfinalcopy 

\def\wacvPaperID{362} 

\ifwacvfinal\fi
\begin{document}

\title{Tukey-Inspired Video Object Segmentation}

\author{Brent A. Griffin ~~~ Jason J. Corso \\
University of Michigan\\
{\tt\small \{griffb,jjcorso\}@umich.edu}
}


\maketitle
\ifwacvfinal\thispagestyle{empty}\fi

\begin{abstract}
	We investigate the problem of strictly unsupervised video object segmentation, i.e., the separation of a primary object from background in video without a user-provided object mask or any training on an annotated dataset.   
	We find foreground objects in low-level vision data using a John Tukey-inspired measure of ``outlierness.''
	This Tukey-inspired measure also estimates the reliability of each data source as video characteristics change (e.g., a camera starts moving). 
	The proposed method achieves state-of-the-art results for strictly unsupervised video object segmentation on the challenging DAVIS dataset.
	Finally, we use a variant of the Tukey-inspired measure to combine the output of multiple segmentation methods, including those using supervision during training, runtime, or both.  
	This collectively more robust method of segmentation improves the Jaccard measure of its constituent methods by as much as 28\%.
\end{abstract}

\section{Introduction}
\label{sec:intro}

Video understanding remains a focus area in vision.
Video object segmentation (VOS), a critical sub-problem, supports learning object class models~\cite{OnReVeECCV2014,TaSuYaCVPR2013}, scene parsing~\cite{LiHeCVPR2015,TiLaIJCV2012}, action recognition~\cite{LuXuCoCVPR2015,SoIdShICCV2015,SoIdShCVPR2016}, and video editing applications~\cite{ChChChACMM2012}. 
Despite the utility of VOS, finding general solutions remains hotly in focus, especially in cases without human annotation or other supervision provided for training or inference.
Most unsupervised methods make use of a measurable property, such as salient object motion \cite{FST}, generic object appearance \cite{IaDe10,KEY}, or rigid background elements \cite{WeSz17}.
However, a primary challenge in VOS is the variability of video characteristics: cameras can be static or moving; backgrounds and objects can be rigid or dynamic; objects can leave view, change scale, or become occluded; and unique elements like rippling water cause peculiar visual effects.
In these circumstances, specific data sources become unreliable, degrading VOS performance.  

\begin{figure}[t]
	\centering
	\includegraphics[width=0.475\textwidth]{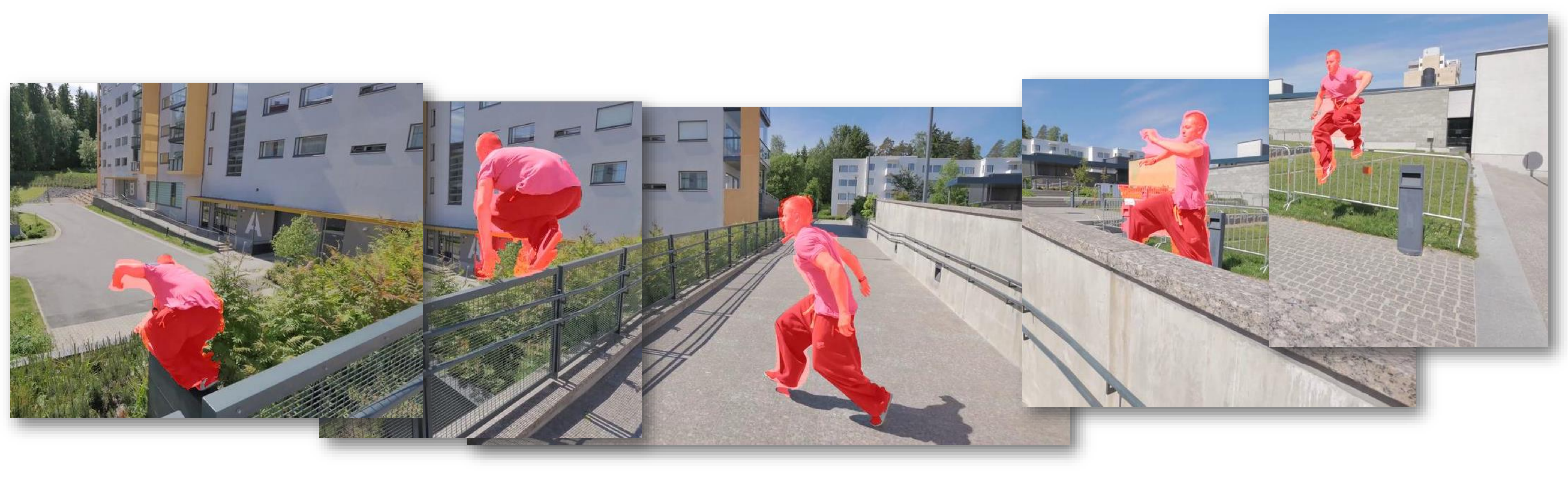}
	\caption{Tukey-inspired segmentation of DAVIS's Parkour video, which has a moving camera, a dynamic foreground object with scale-variation, and occlusions (best viewed in color). 
	}
	\label{fig:parkour}
\end{figure}

We propose a new, strictly unsupervised VOS method called \textbf{T}ukey-\textbf{I}nspired \textbf{S}egmentation (TIS), which separates the primary foreground object in a video from background.
In \textit{Exploratory Data Analysis} \cite{Tukey}, John Tukey provides a statistical method to find outliers in data.
Given that foreground objects typically exhibit a measurable difference relative to the surrounding background, we use Tukey's statistical method to identify candidate foreground objects in low-level vision data.
We also develop our own ``outlierness'' scale to quantitatively determine each data source's ability to reveal foreground objects.
By weighting and combining foreground candidates from multiple data sources, our output segmentation mitigates problems associated with changing video characteristics (see Figure~\ref{fig:parkour}).
In addition, we use a variant of our ``outlierness'' scale to estimate the reliability of segmentations from multiple VOS frameworks, which we weight and combine to generate new, more reliable segmentations. 
Using our TIS method to find foreground objects in low-level vision data, we set a new precedent for unsupervised methods on the DAVIS benchmark dataset.
Using our TIS variant to combine segmentations, we achieve better performance on DAVIS than all prior approaches, whether unsupervised or supervised.

\begin{figure*}[t!]
	\centering
	\includegraphics[width=0.95\textwidth]{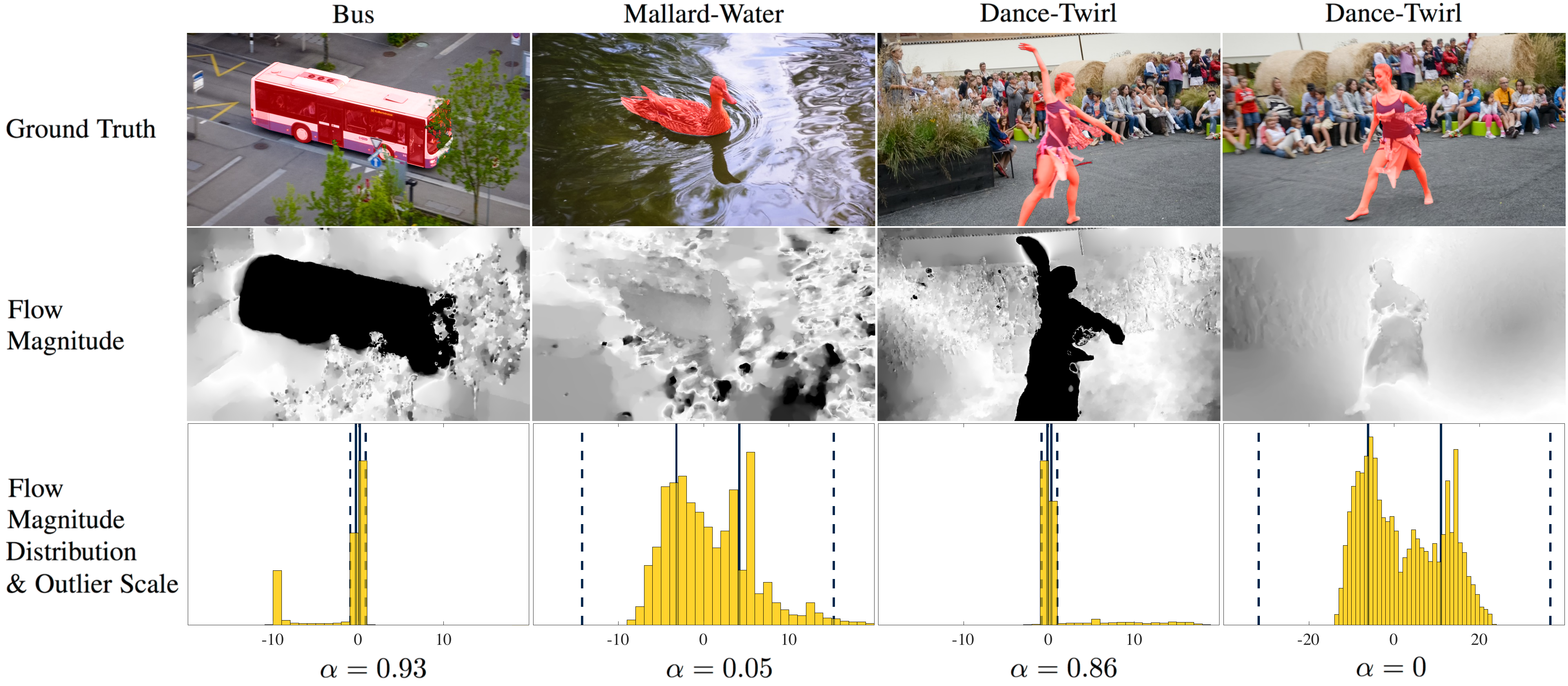}
	\caption{Foreground Candidates in Image Data. The outlier scale $\alpha$ acts as a saliency weighting that adapts to frame-to-frame video characteristics. 
		In these examples, we focus on optical flow magnitude with outliers depicted as black pixels (middle row). 
		Flow distributions are offset from the median (bottom row) and include the interquartile range (solid lines) and outlier thresholds (dotted lines). In the Bus video, the camera follows the bus while the background moves; from the resulting bimodal distribution, we reliably track the bus as a salient flow outlier ($\alpha = 0.93$). In Mallard-Water, due to dynamically flowing background elements, the mallard is difficult to track using flow magnitude ($\alpha = 0.05$). In Dance-Twirl, $\alpha = 0.86 $ when the camera is relatively stationary and $\alpha = 0$ when the camera is moving.}
	\label{fig:saliency}
\end{figure*}

\section{Related Work}

Learning-based methods have become commonplace for VOS benchmarks. 
However, supervised methods require annotated data, which are exceptionally tedious and costly for segmentation.
For example, the state-of-the-art OnAVOS \cite{OnAVOS} has three distinct types of supervision: first, it trains its base network using the ImageNet \cite{imageNet}, Microsoft COCO \cite{MSCOCO}, and PASCAL datasets \cite{PASCAL}; second, it fine-tunes using the DAVIS dataset \cite{DAVIS}; and, third, it requires a user-provided object mask at the beginning of each video.
Other DAVIS benchmark leaders, such as OSVOS-S~\cite{OSVOS-S}, RGMP~\cite{RGMP}, OSVOS~\cite{OSVOS}, and MSK~\cite{MSK}, have similar requirements.
To support learning, some researchers develop their own weakly-annotated (FSEG \cite{FusSeg}) or synthetic training data (LMP \cite{LMP}).
Other methods require supervision through components, such as features from a previously-trained network (OFL \cite{OFL}) or a previously-trained boundary detection algorithm (ARP \cite{ARP,LTDMB}).
Finally, some supervised algorithms are not trained, but still need a user-provided mask at runtime \cite{SEA,JMP,BVS,FCP}.
In contrast, our TIS method has no labeling or training requirements.
This design constraint supports application areas where user-provided masks are impractical and application-specific datasets for training are unavailable. 


Multiple benchmarks are available to evaluate VOS methods \cite{SegTrackv2,SegTrack}, including the Densely Annotated VIdeo Segmentation (DAVIS) dataset \cite{DAVIS}.
DAVIS evaluates VOS methods across many challenge categories, including multiple instances of occlusions, objects leaving view, scale-variation, appearance change, edge ambiguity, camera-shake, interacting objects, and dynamic background (among others); these challenges frequently occur simultaneously.
Using our strictly unsupervised TIS method, we achieve a Jaccard measure (or intersect over union) of 67.6 and, using the combinational variant, a Jaccard measure of 74.9, which is a 17\% improvement over previous unsupervised results on DAVIS \cite{MSG,NLC,TRC,KEY,FST,CVOS,SAL,WeSz17}.

{\color{note}
}

\subsection*{Primary Contributions}

Our primary contribution is a video object segmentation method that utilizes low-level processes without any training or annotation requirements, neither at training or runtime.
Our method achieves the highest performance among all known unsupervised methods and higher performance than many supervised methods on the DAVIS video object segmentation benchmark 
for single objects, which is our focus.
We also develop a variant of our initial approach that adaptively combines segmentations from multiple methods, generating new segmentations that achieve the highest DAVIS performance in every category of unsupervised and supervised approaches.
We provide source code for the current work at \url{https://github.com/griffbr/TIS}.

\section{Tukey-Inspired Segmentation}
\label{sec:approach}

\begin{figure*}[t]
	\centering
	\includegraphics[width=0.975\textwidth]{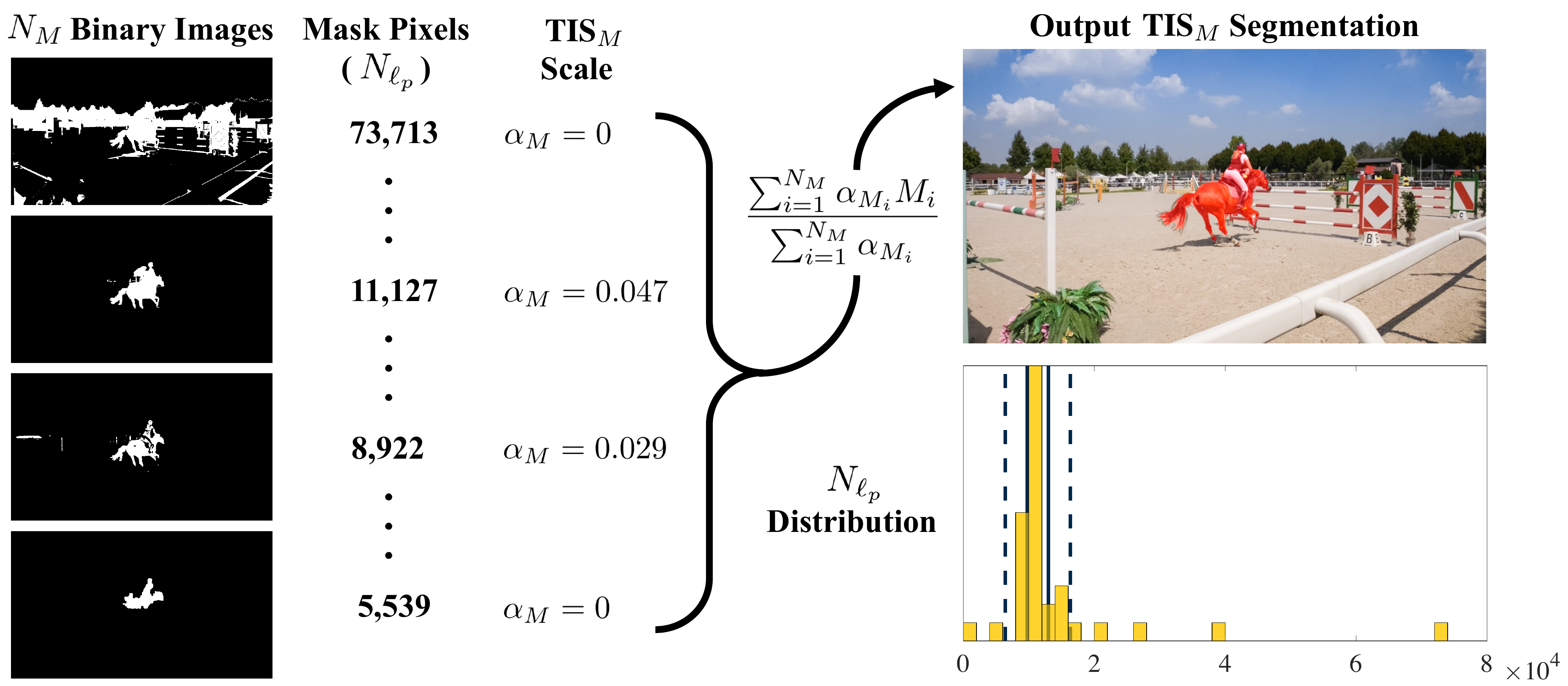}
	\caption{Foreground Objects from Binary Images. 
		In this example, 34 segmentation methods are combined using the TIS$_M$ method (i.e., $N_M=34$).
		The binary image outlier scale $\alpha_M$ functions as a confidence weighting for each mask $M$ and changes each video frame. 
		Masks exhibiting a strong $N_{\ell_{p}}$ consensus with the group are considered reliable (middle left), while masks with too many or too few foreground pixels are considered unreliable outliers (top and bottom left, $\alpha_M=0$).
		The $N_{\ell_{p}}$ distribution for the current video frame (bottom right) includes the interquartile range (solid lines) and outlier thresholds (dotted lines).
	}
	\label{fig:binImg}
\end{figure*}

We derive two Tukey-inspired methods of segmentation. 
In Section~\ref{sec:SO}, we find foreground objects directly from general image data.
We identify foreground candidates using John Tukey's statistical measure of outliers, then we use our own measure of ``outlierness"  to determine the reliability of each data source for identifying foreground objects.
In Section~\ref{sec:sbi}, we combine a set of binary images that represent previous estimates of foreground object locations, which enables the simultaneous utilization of multiple segmentation methods.
To improve accuracy, we use a second ``outlierness" measure to determine the reliability of each source segmentation and weight its relative contribution.

\subsection{Foreground Candidates in Image Data}
\label{sec:SO}

Most foreground objects exhibit a measurable difference in data relative to background elements.
We identify this difference using outliers in vision data (e.g., optical flow).
Given an image $D$ containing data values $d_p\in \mathbb{R}$ for each pixel $p$, we calculate lower and upper outlier thresholds:
\begin{align}
\label{eq:outlierThresh}
{O_1} =& ~{Q_1} - k({Q_3} - {Q_1}) \nonumber \\
{O_3} =& ~{Q_3} + k({Q_3} - {Q_1}),
\end{align}
where ${Q_1}$ and ${Q_3}$ are the lower and upper quartiles for all $d_p\in D$ and ${Q_3} - {Q_1}$ is the interquartile range ($Q_2$ is the median). 
$k$ is a constant that scales the outlier thresholds, ${O_1}$ and ${O_3}$; as suggested by John Tukey in \cite{Tukey}, we use $k=1.5$ to find ``outliers.'' 
We use each set of outlier data
\begin{align}
O := \{ d_p \in D|d_p<{O_1} \vee d_p>{O_3} \}
\label{eq:O}
\end{align}
to identify pixels corresponding to foreground objects. 

We also define our own quantitative measure of ``outlierness" for each data source $D$:
\begin{align}
\alpha := \frac{ \sum_{d_p \in O} |d_p|}{ \sum_{d_p \in D} |d_p| }.
\label{eq:scale}
\end{align}
$\alpha \in [0,1]$ is proportional to the magnitude of data in $O$ relative to $D$.
By calculating $\alpha$ for each data source, we approximate the frame-to-frame capacity of each source to track foreground objects in a variety of video settings (see Figure~\ref{fig:saliency}).
Accordingly, we use $\alpha$ to weight each input data source for our example VOS implementation in Section~\ref{sec:init}. 

\subsection{Foreground Objects from Binary Images}
\label{sec:sbi}
As an alternative to finding foreground objects in image data, we derive a second outlier scale that enables us to combine segmentation masks from multiple methods and weight their relative contribution according to their frame-to-frame reliability (see Figure~\ref{fig:binImg}).
Given a set of $N_M$ binary masks for the same video frame, assume each mask $M$ consists of pixel-level labels, $\ell_p \in \{0,1\}$, where $\ell_{p}=1$ indicates pixel $p$ is a foreground-object location.
Compared to image data, the ``outlierness" of data within each $M$ is relatively meaningless ($\ell_p\in \{0,1\}$ implies that for $\ell_p\in M$:  $Q_1,Q_2,Q_3 \in \{0,1\}$). 
Instead, we measure ``outlierness" across the set of binary images using the total number of foreground pixels in each $M$,
\begin{align}
N_{\ell_p}: = \sum_{\ell_p \in M} \ell_p .
\label{eq:numfg}
\end{align}
Using $N_{\ell_p}$, the outlier scale for each mask is defined as
\begin{align}
\alpha_{M} := \begin{cases}
\text{max}\big(~\frac{N_{\ell_p} - O_1}{Q_2 - O_1}, 0\big) & \text{if} ~N_{\ell_p} < Q_2 \\
\text{max}\big(~\frac{N_{\ell_p} - O_3}{Q_2 - O_3}, 0\big) & \text{otherwise}
\end{cases},
\label{eq:alphaM}
\end{align}
where quartiles $Q_1,Q_2,$ and $Q_3$ and thresholds $O_1$ and $O_3$ are found using \eqref{eq:outlierThresh} on the set of $N_{\ell_p}$ from all $N_M$ masks.

We use $\alpha_{M}\in[0,1]$ to weight each mask based on proximity to the median number of foreground pixels across all $N_M$ masks, with outliers being scaled at 0.
The intuition behind this weighting is simple.
If an individual mask is near the median, it is representative of the collective consensus of segmentation masks for the approximate size of the foreground object, and it is likely more reliable.
Alternatively, if an individual mask is an outlier, it is likely unreliable for the current video frame.

To generate the output segmentation mask, we define an image-level estimate of ``foregroundness" as
\begin{align}
F := \frac{ \sum_{i=1}^{N_M} \alpha_{M_i} M_i }{ \sum_{i=1}^{N_M} \alpha_{M_i} },
\label{eq:TISF}
\end{align}
where $\alpha_{M_i}$ is the outlier scale for the $i$th mask $M_i $. 
Note that $F$'s corresponding pixel-level values $f_{p}\in[0,1]$. 
The final output mask $M$ for the Tukey-inspired combination of binary images is found using
\begin{align}
\ell_{p} = \begin{cases}
1 & \text{if} ~f_{p} >0.5 \\
0 & \text{otherwise}
\end{cases}.
\label{eq:TISSum}
\end{align}

\noindent \textbf{Remark:}
In Figure~\ref{fig:binImg} and Section~\ref{sec:TISMresults}, we refer to this binary image-based segmentation method as TIS$_{M}$.



\section{Tukey-Inspired Segmentation using \\Motion and Visual Saliency Outliers}
\label{sec:init}

We implement the method of finding foreground objects in image data from Section~\ref{sec:SO} using motion and visual saliency data.
For motion saliency, $x$ and $y$ optical flow components are found using the method from \cite{optFlow}; in addition, the flow magnitude (i.e., $|x^2 + y^2|$) and flow angle (i.e., $\arctan(\frac{y}{x})$) are also calculated.
For each flow measure, the outlier thresholds and scales are calculated on a frame-to-frame basis using \eqref{eq:outlierThresh}-\eqref{eq:scale}.
The pixel-level motion saliency measure, $d^{\text{ms}}_{p}$, is defined for each flow component as
\begin{align}
d^{\text{ms}}_{p} := \begin{cases}
0 & \text{if} ~d_p \notin O \vee \alpha_f < 0.5 \\
\alpha_f|d_{p_f} - Q_{2_f}| & \text{otherwise}
\end{cases},
\label{eq:motSal}
\end{align}
where $d_{p_f}$ is the initial pixel-level flow component value with corresponding frame-to-frame median $Q_{2_f}$, outlier scale $\alpha_f$, and a 0.5 minimum scale requirement.

The intuition behind \eqref{eq:motSal} is as follows.
First, whether a foreground object is moving with a fixed camera or vice versa, the foreground object's deviation from the frame's median optical flow will generally be salient (see Bus in Figure~\ref{fig:saliency}).
Second, the absolute value enables a positive ``foregroundness" contribution regardless of a flow component's sign.
Finally, the minimum scale requirement will remove the influence of less reliable flow components.

We found that visual saliency is less useful than optical flow for most video segmentation cases.
However, the product of visual saliency and optical flow is beneficial for videos with dynamic background elements (e.g., Mallard-Water in Figure~\ref{fig:saliency}).
Thus, we include a pixel-level visual saliency measure, $d^{\text{vs}}_p$, defined as
\begin{align}
d^{\text{vs}}_p := ({d_{p_v}})^{k} \sum_{i=1}^{4} \text{max}(\alpha_{f_i},0.5) |d_{p_{f_i}} - Q_{2_{f_i}}|,
\label{eq:visSal}
\end{align}
where $d_{p_v} \in [0,1]$ is a pixel-level visual saliency-based scale (found using \cite{visSal}), $k$ is an exponential scale that adjusts the relative sharpness of $({d_{p_v}})^{k} \in [0,1]$, $d_{p_{f_i}}$ is the $i$th flow component with corresponding median $Q_{2_{f_i}}$ and outlier scale $\alpha_{f_i}$, and the minimum applied scale of 0.5 ensures that visual saliency features are available even if $\alpha_{f_i}=0~ \forall i$.
Three $d^{\text{vs}}_p$ measures are used altogether, with $k=\{1,\frac{1}{2},\frac{1}{3}\}$.

To generate the segmentation mask, we combine the four $d^{\text{ms}}_{p}$ and three $d^{\text{vs}}_p$ saliency measures for a pixel-level estimate of ``foregroundness," defined as
\begin{align}
f_{p} := \sum_{i=1}^{7} d^i_{p},
\label{eq:f0p}
\end{align}
where $d^i_{p}$ is the $i$th measure. 
Using \eqref{eq:f0p}, we generate a mask $M$ for each video frame with pixel-level labels
\begin{align}
\ell_{p} = \begin{cases}
1 & \text{if} ~f_{p} > \beta \delta_p \\
0 & \text{otherwise}
\end{cases},
\label{eq:TIS0}
\end{align}
where $\ell_{p}=1$ indicates a foreground object and $\beta \in \mathbb{R}$ is the sum of the mean and standard deviation of $f_{p}$ in the current frame. 
$\delta_p$ is the pixel-level previous-mask ``discount" 
\begin{align}
\delta_p := \begin{cases}
\frac{1}{2} & \text{if} ~ \ell_{p,{i}-1}=1 \\
1 & \text{otherwise}
\end{cases}.
\label{eq:delta}
\end{align}

In simple words, if $f_{p}$ is greater than the pixel-level mean and standard deviation of ``foregroundness" in the current frame, pixel $p$ is considered a foreground object location. 
In addition, wherever $p$ corresponds to a mask position in the previous frame, a half-threshold discount is applied, which encourages frame-to-frame continuity and gradually increasing accuracy of the segmentation mask.
Finally, the output segmentation mask assumes a single foreground object hypothesis, so the mask in each frame is the single continuous segment with the greatest $f_{p}$ sum.

\noindent \textbf{Remark:}
In the remainder of the paper, we refer to this image data-based segmentation method as TIS$_0$.

	\begin{figure}[t]
	\centering
	\includegraphics[width=0.475\textwidth]{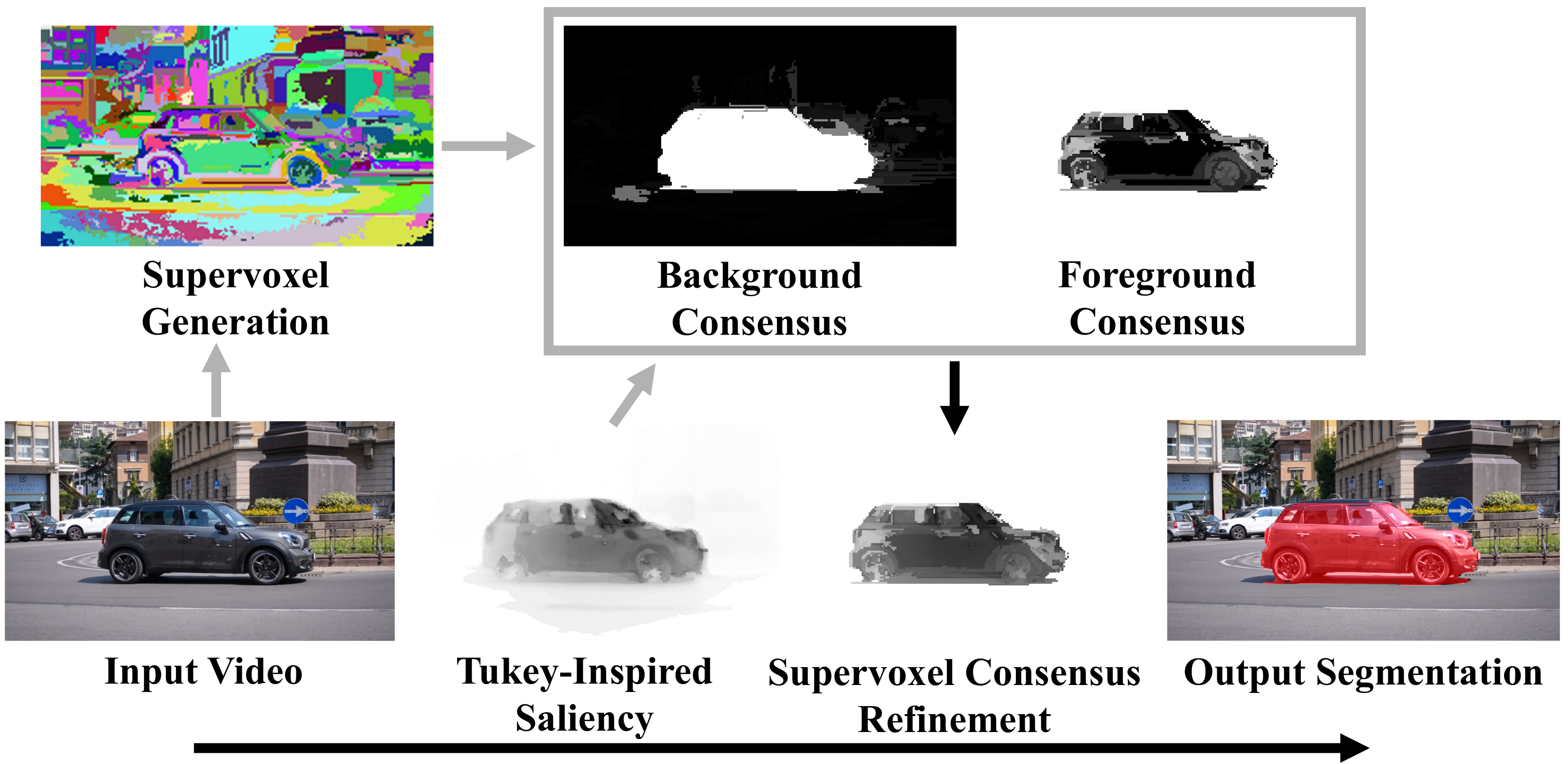}
	\caption{Consensus-based Boundary Refinement. Our Tukey-inspired segmentation  (TIS$_0$) uses saliency outliers in image data to form a quantitative estimate of ``foregroundness" and a corresponding object mask. 
		Using supervoxels generated from the input video, we refine TIS$_0$ with a supervoxel-based consensus for both background and foreground elements, 
		improving the boundary accuracy of the TIS$_0$ segmentation.
	}
	\label{fig:SVXCon}
\end{figure}

\subsection{Consensus-based Boundary Refinement}
\label{sec:SVXCon}

To improve the boundary accuracy of \eqref{eq:TIS0},  we use supervoxel consensus voting.
This choice is motivated by previous VOS work in \cite{NLC} that relates the ``foregroundness'' of superpixels to nearest neighbors across frames.
Supervoxels, on the other hand, inherently exist across many frames and have shown promising results for relating features \cite{binSVX,XuWhCo13} and detecting object boundaries \cite{SVXEval}.
Given a set of non-overlapping supervoxels that cover all video pixels, we use TIS$_0$ to build an internal consensus within the bounds of each supervoxel, and then relate this consensus across the video to refine the boundary of the initial TIS$_0$ mask.
This refinement process is depicted in Figure~\ref{fig:SVXCon}. 
	
	First, using $\ell_{p} \in \{0,1\}$ from \eqref{eq:TIS0}, the local consensus within each supervoxel $S$ is defined as
	\begin{align}
	f^{\text{L}}_{S} := \frac{1}{N_{p}}\sum_{p\in S} (2\ell_{p}-1),
	\label{eq:local}
	\end{align}
	where $N_{p}$ is the number of pixels $p\in S$. Note that $f^{\text{L}}_{S} \in [-1,1]$, where positive or negative values imply a foreground object or background.
	
	Next, the non-local consensus among each $S$ is found as
	\begin{align}
	f^{\text{NL}}_{S} = \sum_{i=1}^{N_{\text{NL}}} w_i f^{\text{L}}_{S_i},
	\label{eq:nonlocal}
	\end{align}
	where $N_{\text{NL}}$ is the number of nearest-neighbors contributing to the non-local consensus for supervoxel $S$ and $w_i$ determines the relative weight of each neighbor.
	Because the total number of supervoxels,  $N_S$, changes with each video, we set $ N_{\text{NL}} = \lceil\frac{N_S}{100}\rceil$.
Nearest neighbor weight $w_i$ is set as 
\begin{align}
w_i = \frac{1}{R(S,S_i)^2},
\label{eq:NNW}
\end{align}
where $R\in \mathbb{R}$ calculates the city-block distance between the mean-LAB color of local supervoxel $S$ and the $i$th nearest neighbor $S_i$.
To ensure that all three LAB distances are meaningful, video-wide LAB values are linearly mapped between 0 and 1.
$R$ is squared to reduce the influence of supervoxels outside of the primary ``clique.''
	
Finally, to refine the TIS$_0$ segmentation, we add \eqref{eq:local} and \eqref{eq:nonlocal} to the initial estimate of ``foregroundness'' from \eqref{eq:f0p}:
	\begin{align}
f'_p := f_{p} + w_0 f^{\text{L}}_{S} + f^{\text{NL}}_{S},
\label{eq:fp}
\end{align}
where $f^{\text{L}}_{S}$ are $f^{\text{NL}}_{S}$ are the local and non-local consensus for the supervoxel containing $p$ and $w_0$ determines the relative weight of $f^{\text{L}}_{S}$ to $f^{\text{NL}}_{S}$.
Essentially, $f'_{p}$ improves the $f_{p}$-based boundary by adding to supervoxels with a consistent foreground object consensus while subtracting from supervoxels with a consistent background consensus. 
	The refined TIS$_0$ segmentation mask is found using pixel-level labels
	\begin{align}
	\ell_{p} = \begin{cases}
	1 & \text{ if } f'_p > 0 \\
	0 & \text{otherwise}
	\end{cases}.
	\label{eq:mask}
	\end{align}
	
	\noindent \textbf{Remarks:}
	(a) $f_{p}$ in \eqref{eq:fp} is scaled s.t. $f_{p} \in [0,1]$.
	(b) If using only local consensus, $w_0=1$ in \eqref{eq:fp} and all non-local weights are zeroed.
	(c) If using local and non-local consensus, $w_0=\frac{1}{3}$ and non-local weights are uniformly scaled $s.t. \sum_{i=1}^{N_{\text{NL}}}w_i=\frac{2}{3}$.
	(d) The refined mask \eqref{eq:mask} is limited to the two segments with the greatest $f'_p$ sum in each frame.
	This improves VOS for objects with partial occlusions.
	
		\setlength{\tabcolsep}{6pt}
	\begin{table}[t]
		\centering
		\caption{DAVIS Results for TIS$_0$ with and without Supervoxel Consensus Refinement. 
			GBH-based refinement performs better when using only the local consensus.
			Higher numbers are better for rows labeled with $\uparrow$ (e.g., $\mathcal{J}$ mean) and worse for rows with $\downarrow$.
		}
		\begin{tabular}{ r | c | c | c}
			\hline
			& \multicolumn{3}{c}{Configuration ID} \\
			\multicolumn{1}{c|}{Configuration} & \multicolumn{1}{c}{TIS$_{\text{S}}$} & \multicolumn{1}{c}{TIS$_{\text{G}}^{\text{L}}$} & \multicolumn{1}{c}{TIS$_0$} \\
			\hline
			Supervoxels Used & SWA & GBH& None   \\
			Hierarchy Level  & 6 & 2& N/A \\
			Local Consensus  & Yes  & Yes& N/A \\
			Non-Local Consensus  & Yes  & No & N/A \\
			\hline
			\multicolumn{1}{c|}{Measure} & \multicolumn{3}{c}{DAVIS Results} \\
			\hline
			Region Similarity: $\mathcal{J}$ Mean $\uparrow$  & \bf{67.6} & 65.3 & 58.6\\
			Contour Accuracy: $\mathcal{F}$ Mean $\uparrow$  & \bf{63.9} & 61.2 & 47.5	 \\
			Temporal Stability: $\mathcal{T}$ Mean $\downarrow$  & 31.0 & 31.8 & \bf{30.7}\\
			\hline
		\end{tabular}
		\label{tab:Config}
	\end{table}

\setlength{\tabcolsep}{9pt}
\begin{table*}[t]
	\centering
	\caption{
		\textbf{Complete DAVIS} Results for State-of-the-Art \textbf{Unsupervised Methods}.
		Methods are unsupervised, so we compare using training and validation videos.
		Object recall measures the fraction of sequences scoring higher than 0.5, and decay quantifies the performance loss (or gain) over time \cite{DAVIS}.
		TIS$_0$ exhibits the best decay performance; TIS$_{\text{S}}$ and TIS$_M$ achieve top results in all other categories. 
	}
	\begin{tabular}{ r | c  c c | c c c c c c c c }
		\hline
		\multicolumn{1}{c |}{} & \multicolumn{3}{c|}{Current Results} & & \cite{WeSz17} \\
		\multicolumn{1}{c |}{Measure} & TIS$_{M}$ & TIS$_{\text{S}}$ &  TIS$_0$ & NLC & BGM & FST & KEY & MSG & CVOS & TRC & SAL  \\
		\hline
		Mean $\uparrow$    &   \bf 74.9  & \bf\ 67.6  &    \ 58.6  &    \ 64.1  &  \ 62.5   &  \ 57.5  &    \ 56.9  &    \ 54.3  &    \ 51.4  &    \ 50.1  &    \ 42.6 \\
		$\mathcal{J}$ Recall $\uparrow$   & \bf 90.1   & \bf\ 84.7  &    \ 75.9  &    \ 73.1  &  \ 70.0   &    \ 65.2  &    \ 67.1  &    \ 63.6  &    \ 58.1  &    \ 56.0 &    \ 38.6 \\
		Decay $\downarrow$   &\ ~5.8  &  \ ~~4.0 &   \bf\ ~2.3  &    \ ~8.6  &  -  &  \ ~4.4  &    \ ~7.5  &    \ ~2.8  &    \ 12.7  &    \  ~5.0 &    \ ~8.4 \\
		\hline
		Mean $\uparrow$      &  \bf 69.0 & \bf\ 63.9  &    \ 47.5  &    \ 59.3  &  \ 59.3  &  \ 53.6  &    \ 50.3  &    \ 52.5  &    \ 49.0 &    \ 47.8  &    \ 38.3 \\
		$\mathcal{F}$ Recall $\uparrow$    &  \bf 83.8 & \bf\ 78.5  &   \ 48.8  &    \ 65.8  &  \ 66.2  &    \ 57.9  &    \ 53.4  &    \ 61.3  &    \ 57.8  &    \ 51.9  &    \ 26.4 \\
		Decay $\downarrow$   & \ ~9.0 &    \ ~~5.7  &    \bf\ ~1.4  &    \ ~8.6  &   -  &   \ ~6.5  &    \ ~7.9  &    \ ~5.7  &    \ 13.8  &    \ ~6.6  &    \ ~7.2 \\
		\hline
	\end{tabular}
	\label{tab:DAVISCompare}
\end{table*}

\setlength{\tabcolsep}{3.95pt}
\begin{table*}[t]
	\centering
	\caption{
		\textbf{DAVIS Validation Set} Results. 
		Runtime supervision requires a user-provided annotation frame for each segmentation video.
		Training supervision includes dataset training and dataset-trained components (e.g., OFL and ARP).
		The online benchmark distinguishes by runtime supervision only. 
		TIS$_0$-based methods achieve top unsupervised results; TIS$_M$-based methods achieve top results in all categories. 
	}
	\begin{tabular}{ r | c c c c c c c | c c | c c c | c c c c c}
		\cline{1-18}
		\multicolumn{1}{c |}{} & \multicolumn{12}{c|}{Supervision Required} \\
		\cline{2-13}
		\multicolumn{1}{c |}{} & \multicolumn{7}{c |}{Runtime \& Training} &  \multicolumn{2}{c |}{Runtime} & \multicolumn{3}{c|}{Training} & \multicolumn{5}{c}{Unsuperivsed} \\
		\cline{2-18}
		\multicolumn{1}{c |}{\tiny{Measure}} & \tiny{TIS$_{M5}^{\text{V}}$} & \tiny{TIS$_{M}^{\text{RTV}}$} & \tiny{OnAVOS} & \tiny{OSVOS-S} & \tiny{CINM} & \tiny{FAVOS} & \tiny{RGMP}& \tiny{TIS$_{M}^{\text{RV}}$}& BVS & \tiny{TIS$_{M}^{\text{TV}}$} & PDB & ARP & \tiny{TIS$_{M}^V$} & TIS$_{\text{S}}$  & TIS$_0$ & FST & CUT \\
		\hline
		$\mathcal{J}$ Mean& \bf 88.1	& \bf 86.7 & 86.1 & 85.6 & 83.4 & 82.4 & 81.5 & \bf 76.5 
		& 60.0 & \bf 81.8 & 77.2 & 76.2 & \bf 71.2 & \bf 62.6 & \bf 56.2 & 55.8 & 55.2\\
		\hline
		$\mathcal{F}$ Mean & \bf 87.5 & 83.8 & 84.9 & \bf 87.5 & 85.0 & 79.5 & 82.0 & \bf 71.2 
		& 58.8 & \bf 76.6 & 74.5 & 70.6 & \bf 66.4 & \bf 59.6 & 45.6 & 51.1 & 55.2 \\
		\hline 
	\end{tabular}
	\label{tab:DAVISVal}
\end{table*}
	
	\section{Results}
	\label{sec:results}
	
	We evaluate our TIS segmentation methods on two VOS benchmark datasets: the Densely Annotated VIdeo Segmentation (DAVIS) dataset \cite{DAVIS} and the Georgia Tech Segmentation and Tracking Dataset (SegTrackv2) \cite{SegTrackv2,SegTrack}.
	The DAVIS 2016 dataset includes 50 diverse videos, 30 training and 20 validation, all of which have ground truth annotations matching the single object hypothesis (unlike DAVIS 2017-18).
	The SegTrackv2 dataset has fewer videos than DAVIS, and only a subset match the single object hypothesis.
	SegTrackv2 also contains videos with different resolutions, which span from 76,800 to 230,400 pixels per frame.
	
	Three standard benchmark measures evaluate the performance of our segmentation method:
	region similarity $\mathcal{J}$, contour accuracy $\mathcal{F}$, and temporal stability $\mathcal{T}$, which are all calculated using the definitions provided in \cite{DAVIS}.
	Region similarity (also known as the intersect over union or Jaccard index \cite{jaccard}) provides an intuitive, scale-invariant evaluation for the number of mislabeled foreground pixels with respect to a ground truth annotation.
	Given a foreground mask $M$ and ground truth annotation $G$, $\mathcal{J}=\frac{M\cap G}{M \cup G}$.
	Contour accuracy evaluates the boundary of a segmentation by measuring differences between the closed set of contours for $M$ and $G$. 
	Finally, temporal stability is a measure based on the consistency of a mask between video frames.

\setlength{\tabcolsep}{9.75pt}
\begin{table*}[t]
	\centering
	\caption{
		\textbf{SegTrackv2} Results.
		Other results are directly from citation or comparative studies in \cite{NLC,FusSeg,YaEtAl16}.
		Videos are single-object only.
	}
	\begin{tabular}{ r | c c c | c c c c c c | c c }
		\hline
		& \multicolumn{9}{c|}{Unsupervised} & \multicolumn{2}{c}{Supervised} \\
		\cline{2-12}
		\multicolumn{1}{c|}{Video} & \multicolumn{3}{c|}{Current Results} &  & \cite{ITS} & \cite{YaEtAl16} & \cite{HPF} &  &  &  &  \\
		\multicolumn{1}{c|}{$\mathcal{J}$ Mean} & TIS$^{\text{L}}_{\text{G}}$ & TIS$_{\text{S}}$  & TIS$_0$ &	NLC & ITS & FAM & HPF & KEY &	FST & FSEG & HVS \\
		\hline
		Birdfall  &   62	 &  54	 &  23	 &  74	 &  73	& 66 &  58	 &  49	 &  18	 &  38	 &  57	\\
		Frog  &   78 &  50  &  61	 &  83	 &  80  & 81 &  58 & 	0   & 	 54 & 	 57 & 	 67	\\
		Girl  &   69 & 	 70 & 	 65 & 	 91 & 	 86 & 82 & 69 & 	 88 & 	 55 & 	 67 & 	 32\\
		Monkey  &   58  & 	 57 & 	 34 & 	 71 & 	 83 & 69 & 69 & 	 79 & 	 65 & 	 8 & 	 62	\\
		Parachute  &   88 &  88 & 	 67 & 	 94 & 	 96 & 90 &	 94 &  96 & 	 76 & 	 52 & 	 69	\\
		Soldier  &    56 & 	 53 & 	 49 & 	 83 & 	 76 & 83 & 6 & 	 67 & 	 4 & 	 7 & 	 67	\\
		Worm  &    77 &  58 & 	 52 & 	 81 & 	 82 &  82 &	 84 &  84 & 	 73 & 	 51 & 	 35	\\
		\hline
		All  &  \bf{70} &  62 & 	 50 & 	 \bf 82 & 	 \bf 82 & \bf 79 &	 \bf 70 & 	 66 & 	 54 & 	 59 & 	 56	\\
		\hline
	\end{tabular}
	\label{tab:SegTrackCompare}
\end{table*}

\subsection{TIS$_0$ Foreground Objects from Image Data}
We evaluate our Tukey-inspired measure for finding foreground objects in image data using the TIS$_0$ implementation from Section~\ref{sec:init}.
To improve boundary accuracy, we refine TIS$_0$ with supervoxel consensus (Section~\ref{sec:SVXCon}) using segmentation by weighted aggregation (SWA) \cite{SWA,SWANature} and hierarchical graph-based (GBH) \cite{GBH} supervoxels (both generated using the LIBSVX library \cite{LIBSVX}).
Both TIS$_0$ refinement configurations are detailed in Table~\ref{tab:Config} with additional analysis provided in \cite{GrCo17}.

DAVIS results for TIS$_0$-based methods are compared with state-of-the-art methods in Tables~\ref{tab:DAVISCompare} and \ref{tab:DAVISVal}.
TIS$_0$ exhibits the best decay performance for $\mathcal{J}$ and $\mathcal{F}$ and achieves a higher $\mathcal{J}$ than all previous unsupervised methods on the DAVIS validation set.
Multiple methods have better contour accuracy than TIS$_0$, but the refinement process from Section~\ref{sec:SVXCon} solves this problem.
TIS$_{\text{S}}$, which improves TIS$_0$ with consensus-based boundary refinement, outperforms all unsupervised segmentation methods for both $\mathcal{J}$ and $\mathcal{F}$.


SegTrackv2 results are provided in Table~\ref{tab:SegTrackCompare}.
We select videos from SegTrackv2 that use a single object hypothesis (like DAVIS).
SegTrackv2 uses videos with different resolutions, causing individual hierarchy levels to have dramatically different supervoxel quantities from one video to the next.
Accordingly, we change supervoxel hierarchy levels for TIS$_{\text{S}}$ and TIS$^{\text{L}}_{\text{G}}$ between videos.
Besides NLC, ITS, and FAM, TIS$^{\text{L}}_{\text{G}}$ achieves top results.


\subsection{TIS$_M$ Foreground Objects from Binary Images}
\label{sec:TISMresults}

\setlength{\tabcolsep}{5.25pt}
\begin{table*}[t]
	\centering
	\caption{DAVIS Results for Multiple TIS$_M$ Configurations. Bold font indicates best performance for a given set of methods combined. 
		With the exception of ``Experimental Sets,'' TIS$_M$ configurations use all applicable methods from the online benchmark (davischallenge.org). 
	}
	\begin{tabular}{ c | l | l | c | c | c}
		\hline
		\multicolumn{4}{c|}{Configuration}  & \multicolumn{2}{c}{Results} \\
		\hline
		\multicolumn{1}{c|}{ID} & \multicolumn{1}{l|}{Supervised} & \multicolumn{1}{l|}{Methods Combined} & Total & $\mathcal{J}$ & $\mathcal{F}$ \\ 
		\hline
		\hline
		\multicolumn{6}{c}{\textbf{Complete DAVIS Dataset}} \\
		\hline
		TIS$_M$ & No & {TIS$_{\text{S}}$, TIS$_0$, FST\cite{FST}, NLC\cite{NLC}, MSG\cite{MSG}, KEY\cite{KEY}, CVOS\cite{CVOS}, TRC\cite{TRC}} & 8 & 74.9 & 69.0 \\
		TIS$_M^{\text{T}}$ & Training & TIS$_M$ set + {ARP\cite{ARP}, FSEG\cite{FusSeg}, LMP\cite{LMP}} & 11 & 80.5 & 74.2 \\
		TIS$_M^{\text{R}}$ & Runtime & TIS$_M$ set + {BVS\cite{BVS}, FCP\cite{FCP}, JMP\cite{JMP}, HVS\cite{GBH}, SEA\cite{SFL}} & 13 & 78.2 & 73.5 \\
		TIS$_M^{\text{RT}}$ & R. \& T. & TIS$_M^{\text{R}}$ $\cup$ TIS$_M^{\text{T}}$ sets  + {MSK\cite{MSK}, CTN\cite{CTN}, VPN\cite{VPN}, OFL\cite{OFL}} & 20 & 83.3 & 78.5 \\
		\hline
		\hline
		\multicolumn{6}{c}{Baseline Comparisons on \textbf{DAVIS Validation Set}} \\
		\hline
		TIS$_M^{\text{V}}$ & No & TIS$_M$ set \& {CUT\cite{CUT}} & 9 & 71.2 & 66.4 \\
		mean$_M$ & No & ``~~~~~~~~~~~~~~~~~~~~~~~~~~~~~~~'' & 9 & \bf 73.3 & \bf 67.7 \\
		median$_M$ & No & ``~~~~~~~~~~~~~~~~~~~~~~~~~~~~~~~'' & 9 & 64.8 & 60.0 \\
		\hline
		TIS$_M^{\text{TV}}$ & Training & TIS$_M^{\text{V}}$ $\cup$ TIS$_M^{\text{T}}$ sets + {PDB\cite{PDB}, LVO\cite{LVO}, SFLU\cite{SFL}} & 15 & \textbf{81.8} & \bf 76.6 \\
		mean$_M$ & Training & ``~~~~~~~~~~~~~~~~~~~~~~~~~~~~~~~~~~~~~~~~~~~~~~~~~~~~~~~~~~~~~~~~~~~~~~~~~~~~~~~'' & 15 & 80.8 & 75.3 \\
		median$_M$ & Training &  ``~~~~~~~~~~~~~~~~~~~~~~~~~~~~~~~~~~~~~~~~~~~~~~~~~~~~~~~~~~~~~~~~~~~~~~~~~~~~~~~'' & 15 & 76.0 & 70.3 \\
		\hline
		TIS$_M^{\text{RV}}$ & Runtime & TIS$_M^{\text{V}}$ $\cup$ TIS$_M^{\text{R}}$ sets  & 14 & \bf 76.5 & \textbf{71.2} \\
		mean$_M$ & Runtime & ``~~~~~~~~~~~~~~~~~~~~~~~~~~~~'' & 14 & \bf 76.7 & 70.8 \\
		median$_M$& Runtime &  ``~~~~~~~~~~~~~~~~~~~~~~~~~~~~''  & 14 & 70.3 & 64.4 \\
		\hline
		TIS$_M^{\text{RTV}}$ & R. \& T. & TIS$_M^{\text{RT}}$ $\cup$ TIS$_M^{\text{TV}}$ sets  + {OSVOS-S\cite{OSVOS-S}, OnAVOS\cite{OnAVOS}, CINM\cite{CINM}, PML\cite{PML},} & 34 & \bf 86.7 & \bf 83.8 \\ & &  {~~~~~~RGMP\cite{RGMP}, FAVOS\cite{FAVOS}, OSVOS\cite{OSVOS}, SFLS\cite{SFL}, OSMN\cite{OSMN}, PLM\cite{PLM}} & & & \\
		mean$_M$ & R. \& T. & ``~~~~~~~~~~~~~~~~~~~~~~~~~~~~~~~~~~~~~~~~~~~~~~~~~~~~~~~~~~~~~~~~~~~~~~~~~~~~~~~~~~~~~~~~~~~~~~~~~~~~~~~~~~~~~~~~'' & 34 & 86.1 & 82.9 \\
		median$_M$ & R. \& T. & ``~~~~~~~~~~~~~~~~~~~~~~~~~~~~~~~~~~~~~~~~~~~~~~~~~~~~~~~~~~~~~~~~~~~~~~~~~~~~~~~~~~~~~~~~~~~~~~~~~~~~~~~~~~~~~~~~''  & 34 & 80.6 & 77.1 \\
		\hline
		\hline
		\multicolumn{6}{c}{Experimental Sets on \textbf{DAVIS Validation Set}} \\
		\hline
		TIS$_{M5}^{\text{V}}$ & R. \& T. & {OSVOS-S, OnAVOS, CINM, RGMP, FAVOS} & 5 & 88.1 & \bf 87.5 \\
		mean$_M$ & R. \& T. &  ``~~~~~~~~~~~~~~~~~~~~~~~~~~~~~~~~~~~~~~~~~~~~~~~~~~~~~~~~~~~~~~~~~~~~~~'' & 5 & \bf 88.5 & \bf 87.7 \\
		\hline
		TIS$_{M19}^{\text{V}}$ & R. \& T. & TIS$_{M5}^{\text{RTV}}$  and TIS$_M^{\text{RV}}$ sets & 19 & \bf 85.8 & \bf 82.6 \\
		mean$_M$ & R. \& T. &  ``~~~~~~~~~~~~~~~~~~~~~~~~~~~~~~~~~''& 19 & 83.8 & 80.4 \\
		\hline
	\end{tabular}
	\label{tab:TISM}
\end{table*}

We evaluate our Tukey-inspired measure for finding foreground objects in binary images (Section~\ref{sec:sbi}) by combining the segmentation methods listed in Table~\ref{tab:TISM}.
Each combination set includes a certain level of supervision: strictly unsupervised (TIS$_M$), training on an annotated dataset (TIS$_M^{\text{T}}$), user-provided object mask at runtime (TIS$_M^{\text{R}}$),  or both user-provided masks and dataset training (TIS$_M^{\text{RT}}$).
Some source segmentation methods are only available on the DAVIS Validation Set; TIS$_M$ configurations using these validation methods are only evaluated on the validation set and are distinguished by a V (e.g., TIS$_M^{\text{V}}$).

\begin{figure*}[t!]
	\centering
%
%
\begin{tikzpicture}
\definecolor{mycolor5}{rgb}{0.486,0.13,0.12}
\definecolor{mycolor4}{rgb}{0.545,0.588,0.318}
\definecolor{mycolor2}{rgb}{0.00000,0.15290,0.29800}
\definecolor{mycolor3}{rgb}{1.00000,0.79610,0.01961}
\definecolor{mycolor1}{rgb}{0.00000,0.15290,0.29800}
\begin{axis}[%
width=11.25cm,
height=9.25cm,
at={(2.6in,1.149271in)},
scale only axis,
clip=false,
xmin=0.45,
xmax=0.85,
xlabel={Contour Accuracy ($\mathcal{F}$)},
xtick={0.4,0.5,0.6,0.7,0.8},
xmajorgrids,
ymin=0.5,
ymax=0.85,
ytick={0.4,0.5,0.6,0.7,0.8},
ylabel={Region Similarity ($\mathcal{J}$)},
ymajorgrids,
grid style={black!3},
minor tick num = 1,
legend style={at={(0.025,0.975)},anchor=north west},
legend cell align=left,
axis x line*=bottom,
axis y line*=left
]
\addplot [color=mycolor1,only marks,mark=*,mark options={solid}]
  table[row sep=crcr]{%
0.593	0.641\\
};
\addlegendentry{Unsuperivsed};
\node[right, align=left, inner sep=0mm, text=black]
at (axis cs:0.597,0.639,0) {NLC};

\addplot [color=mycolor1,only marks,mark=*,mark options={solid},forget plot]
table[row sep=crcr]{%
	0.593	0.625\\
};
\node[right, align=left, inner sep=0mm, text=black]
at (axis cs:0.598,0.625,0) {BGM};

\addplot [color=mycolor1,only marks,mark=*,mark options={solid},forget plot]
table[row sep=crcr]{%
	0.536	0.575\\
};
\node[right, align=left, inner sep=0mm, text=black]
at (axis cs:0.541,0.575,0) {FST};

\addplot [color=mycolor1,only marks,mark=*,mark options={solid},forget plot]
table[row sep=crcr]{%
	0.503	0.569\\
};
\node[right, align=left, inner sep=0mm, text=black]
at (axis cs:0.506,0.565,0) {KEY};

\addplot [color=mycolor1,only marks,mark=*,mark options={solid},forget plot]
table[row sep=crcr]{%
	0.525	0.543\\
};
\node[right, align=left, inner sep=0mm, text=black]
at (axis cs:0.53,0.543,0) {MSG};

\addplot [color=mycolor1,only marks,mark=*,mark options={solid},forget plot]
table[row sep=crcr]{%
	0.49	0.514\\
};
\node[right, align=left, inner sep=0mm, text=black]
at (axis cs:0.495,0.515,0) {CVOS};

\addplot [color=mycolor1,only marks,mark=*,mark options={solid},forget plot]
table[row sep=crcr]{%
	0.478	0.501\\
};
\node[right, align=left, inner sep=0mm, text=black]
at (axis cs:0.457,0.511,0) { TRC};

\addplot [color=mycolor2,only marks,mark=*,mark options={solid},forget plot]
  table[row sep=crcr]{%
0.639 0.676\\
};
\node[right, align=left, inner sep=0mm, text=black]
at (axis cs:0.643,0.678,0) { TIS$_{\text{S}}$};

\addplot [color=mycolor2,only marks,mark=*,mark options={solid},forget plot]
table[row sep=crcr]{%
	0.612 0.653\\
};
\node[right, align=left, inner sep=0mm, text=black]
at (axis cs:0.614,0.659,0) { TIS$^{\text{L}}_{\text{G}}$};

\addplot [color=mycolor2,only marks,mark=*,mark options={solid},forget plot]
table[row sep=crcr]{%
	0.475 0.586\\
};
\node[right, align=left, inner sep=0mm, text=black]
at (axis cs:0.48,0.586,0) { TIS$_0$};




\addplot [color=mycolor2,only marks,mark=*,mark options={solid},forget plot]
table[row sep=crcr]{%
	0.69 0.749\\
};
\node[right, align=left, inner sep=0mm, text=black]
at (axis cs:0.695,0.741,0) { TIS$_{M}$};

\addplot [color=mycolor3,only marks,mark=*,mark options={solid}]
table[row sep=crcr]{%
	0.711	0.763\\
};
\addlegendentry{Supervised Training};
\node[right, align=left, inner sep=0mm, text=black]
at (axis cs:0.70,0.773,0) { ARP};

\addplot [color=mycolor3,only marks,mark=*,mark options={solid},forget plot]
table[row sep=crcr]{%
	0.658 0.716\\
};
\node[right, align=left, inner sep=0mm, text=black]
at (axis cs:0.66,0.724,0) { FSEG};

\addplot [color=mycolor3,only marks,mark=*,mark options={solid},forget plot]
table[row sep=crcr]{%
	0.663 0.697\\
};
\node[right, align=left, inner sep=0mm, text=black]
at (axis cs:0.668,0.697,0) { LMP};

\addplot [color=mycolor3,only marks,mark=*,mark options={solid},forget plot]
table[row sep=crcr]{%
	0.742 0.805\\
};
\node[right, align=left, inner sep=0mm, text=black]
at (axis cs:0.732,0.817,0) { TIS$^{\text{T}}_{M}$};

\addplot [color=mycolor4,only marks,mark=*,mark options={solid}]
table[row sep=crcr]{%
	0.656	0.665\\
};
\addlegendentry{Supervised Runtime};
\node[right, align=left, inner sep=0mm, text=black]
at (axis cs:0.661,0.663,0) { BVS};

\addplot [color=mycolor4,only marks,mark=*,mark options={solid},forget plot]
table[row sep=crcr]{%
	0.735 0.782\\
};
\node[right, align=left, inner sep=0mm, text=black]
at (axis cs:0.74,0.782,0) { TIS$^{\text{R}}_{M}$};

\addplot [color=mycolor4,only marks,mark=*,mark options={solid},forget plot]
table[row sep=crcr]{%
	0.546 0.631\\
};
\node[right, align=left, inner sep=0mm, text=black]
at (axis cs:0.551,0.631,0) { FCP};

\addplot [color=mycolor4,only marks,mark=*,mark options={solid},forget plot]
table[row sep=crcr]{%
	0.586 0.607\\
};
\node[right, align=left, inner sep=0mm, text=black]
at (axis cs:0.591,0.607,0) { JMP};

\addplot [color=mycolor4,only marks,mark=*,mark options={solid},forget plot]
table[row sep=crcr]{%
	0.576 0.596\\
};
\node[right, align=left, inner sep=0mm, text=black]
at (axis cs:0.581,0.589,0) { HVS};

\addplot [color=mycolor5,only marks,mark=*,mark options={solid}]
table[row sep=crcr]{%
	0.758	0.803\\
};
\addlegendentry{Supervised Runtime \& Training};
\node[right, align=left, inner sep=0mm, text=black]
at (axis cs:0.763,0.803,0) { MSK};

\addplot [color=mycolor5,only marks,mark=*,mark options={solid},forget plot]
table[row sep=crcr]{%
	0.78 0.832\\
};
\node[right, align=left, inner sep=0mm, text=black]
at (axis cs:0.785,0.833,0) { TIS$^{\text{RT}}_{M}$};

\addplot [color=mycolor5,only marks,mark=*,mark options={solid},forget plot]
table[row sep=crcr]{%
	0.714 0.755\\
};
\node[right, align=left, inner sep=0mm, text=black]
at (axis cs:0.718,0.76,0) { CTN};

\addplot [color=mycolor5,only marks,mark=*,mark options={solid},forget plot]
table[row sep=crcr]{%
	0.724 0.75\\
};
\node[right, align=left, inner sep=0mm, text=black]
at (axis cs:0.729,0.747,0) { VPN};

\addplot [color=mycolor5,only marks,mark=*,mark options={solid},forget plot]
table[row sep=crcr]{%
	0.679 0.711\\
};
\node[right, align=left, inner sep=0mm, text=black]
at (axis cs:0.684,0.711,0) { OFL};

\end{axis}
\end{tikzpicture}%
	\caption{\textbf{Visual Comparison} of Segmentation Methods on \textbf{Complete DAVIS Dataset}. Note the correlation between $\mathcal{J}$ and  $\mathcal{F}$. The TIS$_M$ segmentation methods improve performance across all categories of supervised and unsupervised methods.
	}
	\label{fig:DAVISCompare}
\end{figure*}
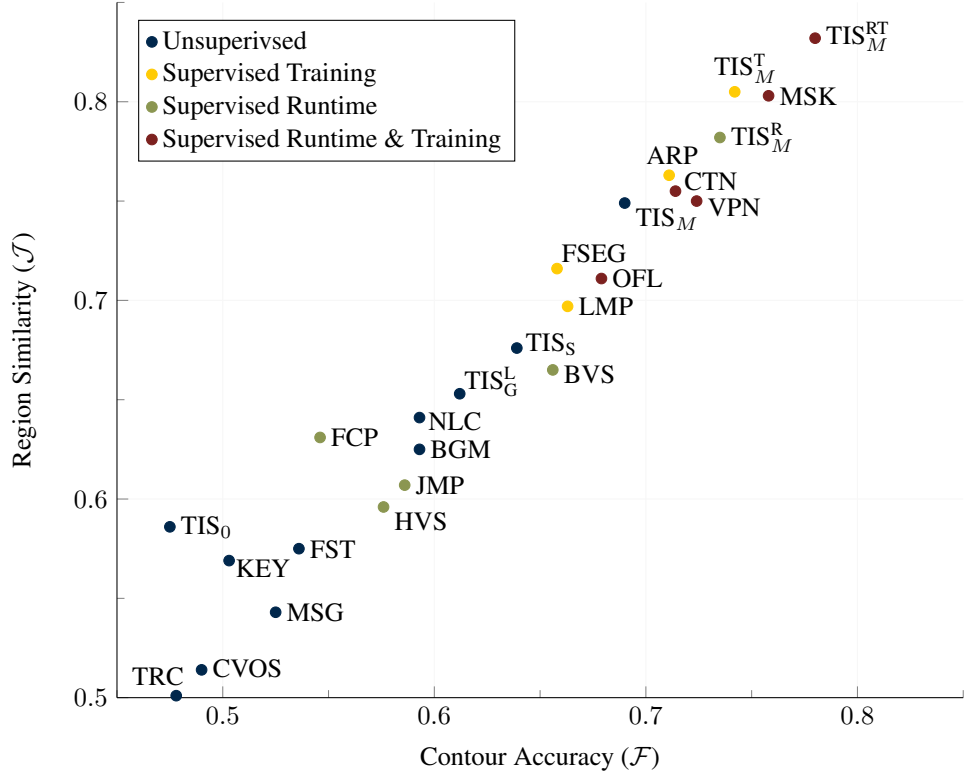

TIS$_M$-based methods achieve top DAVIS results for unsupervised segmentation and all categories of supervised segmentation, as shown in Table~\ref{tab:DAVISVal} and Figure~\ref{fig:DAVISCompare}.
For the complete dataset, the relative $\mathcal{J}$ increases over previous results in each category are: 17\% for unsupervised (TIS$_M$ to NLC), 6\% for training (TIS$_{M}^{\text{T}}$ to ARP), 18\% for runtime (TIS$_{M}^{\text{R}}$ to BVS), and 4\% for runtime and training (TIS$_{M}^{\text{RT}}$ to MSK).
For the validation set, the relative increases are: 28\% for unsupervised (TIS$_{M}^{\text{V}}$ to FST), 6\% for training (TIS$_{M}^{\text{TV}}$ to PDB), 28\% for runtime (TIS$_{M}^{\text{RV}}$ to BVS), and 2\% for runtime and training (TIS$_{M5}^{\text{V}}$ to OnAVOS).
The greatest increases for TIS$_M$ combinations over constituent methods occur for categories with a lower $\mathcal{J}$ score (see Figure~\ref{fig:DAVISCompare}).

To further evaluate TIS$_M$, we compare against two additional statistics-based methods for combining segmentations.
First, we test a mean-based combination (mean$_M$) that averages all source masks together and outputs a foreground label where the pixel-level mean is higher than 0.5; this is equivalent
to setting $\alpha_{M_i} = 1$ in \eqref{eq:TISF}.
Second, we test a median-based combination (median$_M$) that outputs the source mask with the median number of foreground pixels ($N_{\ell_{p}}$ in \eqref{eq:numfg}).
For the baseline comparison in Table~\ref{tab:TISM}, TIS$_M$ has the highest combined $\mathcal{J}$ and $\mathcal{F}$ score for all categories of supervision but one (mean$_M$ over TIS$_M^{\text{V}}$).
In addition, TIS$_M$ is the only combination method that achieves higher $\mathcal{J}$ scores than all source segmentation methods. 

As a final experiment for combining segmentations, we additionally test a small set of the top-five methods (TIS$_{M5}^{\text{V}}$) and a larger set combining the same top-five methods with many poorer segmentations (TIS$_{M19}^{\text{V}}$).
We postulate that the TIS$_M$ combination method is well-suited for the larger, more variable set of segmentations, where the elimination of poorer-performing outliers and promotion of reliable inliers is critical.
This is evidenced by TIS$_{M19}^{\text{V}}$ and the other large TIS$_M$ sets in Table~\ref{tab:TISM} outperforming mean$_M$.
On the other hand, on a smaller, less variable set of segmentation methods, the calculation of quartiles and outliers for scaling is less meaningful; consequently, we find that the small-set TIS$_{M5}^{\text{V}}$ and mean$_M$ have similar performance.




\section{Discussion and Future Work}
\label{sec:conclude}


We develop a Tukey-inspired segmentation methodology to discover foreground objects in general image data.
Our approach automatically adjusts its reliance on each data source according to the frame-to-frame characteristics of a video, addressing a primary challenge in video object segmentation of maintaining reliable measures with changing video characteristics.
In addition, our Tukey-inspired measure for finding foreground objects in image data sets a new precedent on the DAVIS dataset for unsupervised segmentation methods.
Our current implementation used optical flow and visual saliency data, but the method can incorporate additional sources of image data to accommodate new applications.


We apply a variant of our Tukey-inspired measure to combine the output of other segmentation methods, generating a new and collectively more robust method of segmentation.
Our combination method achieves better performance on the DAVIS dataset than all prior segmentation methods and represents a new paradigm in video object segmentation.
In real-world applications, it is difficult to know which individual segmentation methods will perform best for various videos.
On the other hand, by using our approach, multiple methods can be implemented simultaneously, and only the most reliable methods will be used for segmentation in any given video frame.
This extension was particularly effective when constitute methods had variable performance and a large gap for improvement, indicating that the Tukey-inspired combination can be a viable tool for cutting-edge applications where performance has not yet reached its potential.
Furthermore, we found that attempting to combine segmentations using ``non-Tukey'' methods can result in worse performance than some of the source segmentations.
Finally, our combination method can easily incorporate new, better-performing segmentation methods as they are developed by the research community.


%


Given that the current results are restricted to a single object hypothesis, we are currently working on extending this approach to multiple objects.
We are also exploring how our method of weighting input data based on ``outlierness" can improve performance in supervised learning applications.


%
%
%

\section*{Acknowledgements}
\noindent This work was partially supported by the DARPA MediFor program under contract FA8750-16-C-0168.

{\small
\bibliographystyle{ieee}
\bibliography{GrCoWACV19}
}

\end{document}